\pdfoutput=1

\documentclass[11pt]{article}
\usepackage{acl}
\usepackage{times}
\usepackage{latexsym}
\usepackage{graphicx}
\usepackage[T1]{fontenc}
\usepackage[utf8]{inputenc}
\usepackage{microtype}
\usepackage{cleveref}
\usepackage{booktabs}
\usepackage{multirow}
\usepackage{amssymb}

\newcommand{\thick}{\specialrule{.1em}{.05em}{.05em}}

\newcommand{\lengths}{$L$=\color{blue}{128} & $L$=\color{blue}{256} & $L$=\color{blue}{512}}
    
\newcommand{\longresult}[3]{#1 & \multicolumn{3}{c}{#2\,({\color{blue}1024})\quad #3\,({\color{blue}$\infty$})}}

\title{The NLP Task Effectiveness of Long-Range Transformers}

\author{
Guanghui Qin \quad Yukun Feng \quad Benjamin Van Durme \\
Department of Computer Science, Johns Hopkins University \\
{\tt \{gqin2, yfeng55, vandurme\}@jhu.edu }
}

\begin{document}
\maketitle

\begin{abstract}
Transformer models cannot easily scale to long sequences due to their $O(N^2)$ time and space complexity.
This has led to Transformer variants seeking to lower computational complexity,
such as Longformer and Performer.
While such models have theoretically greater efficiency,
their effectiveness on real NLP tasks has not been well studied.
We benchmark 7 variants of Transformer models on 5 difficult NLP tasks and 7 datasets.
We design experiments to isolate the effect of pretraining and hyperparameter settings, to focus on their capacity for long-range attention.
Moreover, we present various methods to investigate attention behaviors
to illuminate model details  beyond  metric scores.
We find that the modified attention in long-range transformers has advantages on content selection and query-guided decoding, but they come with previously unrecognized drawbacks such as insufficient attention to distant tokens and accumulated approximation error. 
\end{abstract}

\section{Introduction}
\vspace{-4pt}

Transformer-based models~\citep{transformers2017} have advanced the state of the art in natural language processing.
However, their quadratic time and space complexity hinder their application on long texts.
Various proposals have been made to address these concerns~\citep{tay2020EfficientTransformersSurvey}, with mathematical guarantees on improved time or space. These  models have been evaluated primarily via perplexity~\citep{dai2019TransformerXLAttentiveLanguage} and non-NLP benchmarks~\citep{tay2020LongRangeArena}. These metrics may not be ideal~\citep{sun2021LongRangeLanguageModels} and may not reflect performance on complex NLP tasks~\citep{arutiunian2020ReproducibilityChallengeReformer,thorne2021DatabaseReasoningText}.
We argue these metrics have not been sufficient for the development of efficient Transformers and their practical application on long texts,
and that existing benchmarks are insufficient guides for architecture selection.

It is not straightforward to have a fair and side-by-side comparison among those models due to the differences between their pretraining and hyperparameter settings~\citep{tay2020EfficientTransformersSurvey},
and the metrics alone cannot convey detailed information 
about the self-attention blocks~\citep{sun2021LongRangeLanguageModels}.
We wish to fairly validate the effectiveness of long-range attention techniques,
and to uncover the underlying factors behind model behaviors.
We critique the reliance on perplexity evaluations in previous work, experimenting with five difficult, long-text NLP tasks.
These tasks cover typical NLP modeling scenarios: token or span-level prediction, sequence-level classification, and sequence-to-sequence generation. 
To our knowledge, 
this is the first work to evaluate long-range transformers on such a wide spectrum of representative NLP tasks.

To verify the key features of long-range transformers,
we \textit{ablate} distant attention to measure what they gain from long-range mechanisms.
For models without pretrained checkpoints, we migrate parameters from their prototype models for fairness.
We cover 3 main kinds of long-range transformers, including pattern-, recurrence-, and kernel-based methods.
To our knowledge, we are the first to adopt all these methods to probe transformers. %
To investigate the relationship between performance and document lengths
we break down the metric with a customized algorithm~\citep{bagga1998AlgorithmsScoringCoreference}.
Also, we use entropy and attribution analysis~\cite{li2017VisualizingUnderstandingNeural} to test the effectiveness of cached memories in recurrent transformers and the global tokens for query-based problems.

We find that long-range context brings performance gains to transformers in some cases,  which we attribute to more selective attention,
especially for query-based tasks like QA.
Surprisingly we observe that some long-range models do not effectively utilize distant information,
and the accumulated error of approximation is unacceptable.
We hope this analysis helps practitioners better understand the current state of the art of long-range attentions and suggests paths for future research.

\begin{figure*}[h!]
    \centering
    \includegraphics[scale=0.6]{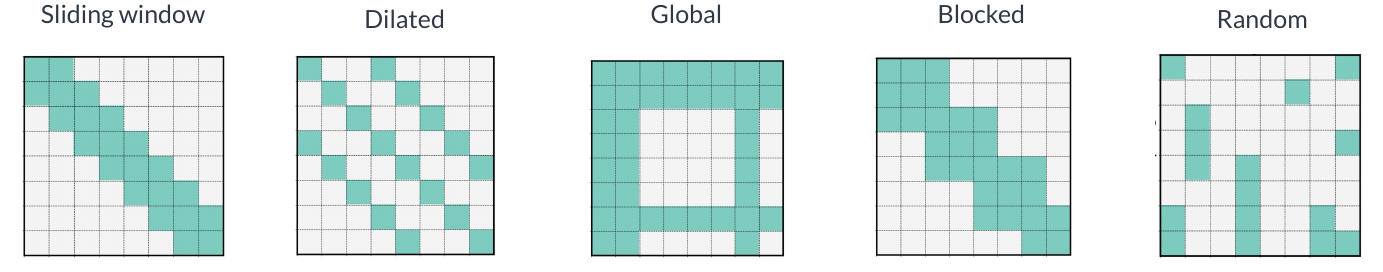}
    \caption{Illustration of 5 patterns used by long-range transformers, from \citet{beltagy2021ParagraphsNLPLong} with permission.}
    \label{fig:pattern_diagram}
\end{figure*}

\section{Background}

\subsection{Long-Range Transformers}
\label{sec:long-transformers}

Researchers have proposed a number of Transformer variants~\citep{tay2020EfficientTransformersSurvey}.
Most of these models support decoding (causal masking)~\citep{peng2021ABCAttentionBoundedmemory}, while only a few of them have pretrained checkpoints~\citep[inter alia]{beltagy2020LongformerLongDocumentTransformer}.
We cluster these approaches into 3 main categories. %

\paragraph{Sparsified Patterns}
Pattern-based methods try to make self-attention sparse.
Some apply pre-specified attention patterns.
Specifically, Longformer~\citep{beltagy2020LongformerLongDocumentTransformer} applies 3 patterns: 
\emph{Sliding window} requires that each token can only attend to the tokens in a local window, 
\emph{dilated pattern} lets each token only attend at fixed intervals, 
while the \emph{global pattern} requires a few tokens as globally attended and lets them to attend all tokens in the sequence. 
In addition to the global pattern, BigBird~\citep{zaheer2020BigBirdTransformers} applies a \emph{blocked pattern}, which splits the sequence into fixed-length blocks, and \emph{random patterns}, by which tokens can attend to any other tokens randomly.
An illustration is shown in \cref{fig:pattern_diagram}.
Although the attention of each layer is not full,
the receptive field can be increased as multiple layers are stacked.
The selected or appended ``global'' tokens can be task-specific~\citep{beltagy2020LongformerLongDocumentTransformer},
allowing for direct distant information exchange.
Instead of pre-defined attention patterns,
some use content-based patterns so they become learnable, with techniques including locality sensitive hashing~\citep{kitaev2020ReformerEfficientTransformer}, the differentiable Sinkhorn algorithm~\citep{tay2020SparseSinkhornAttention}, or the learnable routing algorithm~\citep{roy2020EfficientContentBasedSparse}.

\begin{figure}[t]
    \centering
    \includegraphics[scale=1.2]{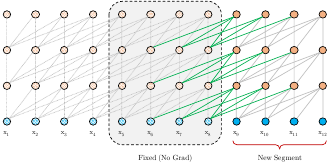}
    \caption{Recurrent transformers. ``No Grad'' means that the gradients do not back-propagate to this block.
    Obtained from \citet{dai2019TransformerXLAttentiveLanguage} with permission.
    }
    \label{fig:recurrence}
\end{figure}

\paragraph{Recurrence \& Compressed Memory}
These methods use segment-level recurrence to reuse the cached hidden states of previous steps.
Transformer-XL~\citep{dai2019TransformerXLAttentiveLanguage} and XLNet~\citep{yang2019XLNetGeneralizedAutoregressive} connects different chunks with cross-attention, where the tokens in a block attend to the hidden states of the previous blocks in addition to their self-attention.
Note that the gradients remain in the same segment and are not propagated to previous segments (\cref{fig:recurrence}).
To reduce the number of history hidden states, \citet{compressive2020} compress them as memories for efficient re-use with pooling or convolutions.
From a different perspective,
\citet{izacard2021LeveragingPassageRetrieval} use retrieval-based methods to collect evidences from external knowledge,
resulting in a more targeted context information.

\paragraph{Low-Rank \& Kernels}
These methods approximate the self-attention with low-rank approximation \citep{wang2020LinformerSelfAttentionLinear} or kernelization
without explictly computing the matrix production.
Among them, \citet{choromanski2021RethinkingAttentionPerformers,peng2021RandomFeatureAttention} use random features.
\citet{katharopoulos2020TransformersAreRNNs} reduce the time complexity to linear and space complexity to constant by replacing softmax with linear kernel features.

\subsection{Benchmarks and Analysis}

There is not an agreed-upon standard benchmark for long-range transformers.
Researchers have considered various tasks and domains,
including language~\citep{dai2019TransformerXLAttentiveLanguage}, protein sequences~\citep{choromanski2021RethinkingAttentionPerformers}, and images~\citep{katharopoulos2020TransformersAreRNNs}.
Few conduct experiments on long sequence NLP tasks, 
including question answering~\citep{beltagy2020LongformerLongDocumentTransformer} and summarization~\citep{zaheer2020BigBirdTransformers}.
\citet{tay2020LongRangeArena} propose Long Range Arena comprised of six non-NLP tasks to exclude the factor of pretraining.
Concurrent to our efforts, \citet{shaham2022SCROLLSStandardizedCompaRison} propose a suite of text-to-text tasks as a long sequence NLP benchmark.

Researchers are also interested in the utility of context in transformers.
\citet{rae2020TransformersNeedDeep} find that Transformer-XL does not necessarily need long and deep contexts.
\citet{sun2021LongRangeLanguageModels} reveal that Longformer and Routing transformers can only reduce the perplexity of LMs on a small set of tokens.
More related to our work, \citet{lai2020ContextAnalysisPretrained} show that BERT can make use of a larger scope of context than a BiLSTM.

\section{Setup}

\subsection{Settings}
It is non-trivial to  compare distinct transformer models, due to differences between their pretraining and hyper-parameter settings.
Our goal is to minimize these confounding factors to allow a focus on the long range attention ability of each model on different tasks. We therefore propose two sets of experimental conditions.
\vspace{-3pt}
\paragraph{Restricted Attention Range}
To evaluate the performance gain from long-range attention, we evaluate models in both their default \textit{context-aware} settings
and a \textit{context-agnostic} setting that restricts the receptive range of the self-attention blocks.
For pattern-based transformers, we achieve the restriction by segmenting the input sequence into chunks and running the transformers on segments independently.
For recurrent models, we ablate the recurrence to eliminate the dependencies between segments.
In practice, we segment the texts into chunks with length $L$,\footnote{We use wordpieces instead of words in this paper.} which ranges from 128 to 1536. $L$=$\infty$ indicates no segmentation is used.

\vspace{-3pt}
\paragraph{Parameter Migration}
Kernel-based models usually do not come with checkpoints for general tasks,
but they may have similar structures to other pretrained models like BERT and may be designed to approximate the original results.
Therefore, it is feasible to migrate parameters from pretrained prototypes to their ``efficient version'' to observe if the performance could be preserved.
This type of method can be  suitable for models without additional parameters, such as Performer.

\subsection{Transformers and Tasks}

\vspace{-3pt}
\paragraph{Transformers} We consider three approaches: 1) {\it pattern-based}: Longformer~\citep{beltagy2020LongformerLongDocumentTransformer}, and BigBird~\citep{zaheer2020BigBirdTransformers};
2) {\it recurrent}: XLNet~\citep{yang2019XLNetGeneralizedAutoregressive}; 
and 3) {\it kernel-based}:  Performer~\citep{choromanski2021RethinkingAttentionPerformers}.
We also include the results of RoBERTa~\citep{roberta2019} and SpanBERT~\citep{joshi2020SpanBERTImprovingPretraining}, where some of our approaches are initialized from those two non-long-range models.
Due to  memory requirements,
we use the base version for all models.\footnote{We used the codebase of \citet{katharopoulos2020TransformersAreRNNs} for Performer and Huggingface~\citep{wolf2020TransformersStateoftheArtNatural} for the rest.}
You may refer to \cref{sec:app_encoder} for more details.

\vspace{-3pt}
\paragraph{Tasks}
We cover five tasks of three types,
including 1) {\it span-level} predictions: coreference resolution (Coref.)~\citep{weischedel2011OntoNotesLargeTraining} and extractive question answering (eQA)~\citep{joshi2017TriviaQALargeScale};
2) {\it sequence classification}: natural language inference (NLI)~\citep{yin2021DocNLILargescaleDataset};\footnote{DocNLI is modified from  ANLI~\citep{nie2020AdversarialNLINew}, SQuAD~\citep{rajpurkar2016SQuAD100000}, DUC2001, DailyMail~\citep{nallapati2016AbstractiveTextSummarization}, and Curation~\citep{curation2020CurationCorpusBase}}
and 3) {\it seq2seq}: summarization (Summ.)~\citep{chen2021SummScreenDatasetAbstractive,huang2021EfficientAttentionsLong}, abstractive QA (aQA)~\citep{dasigi2021DatasetInformationSeekingQuestions,pang2022QuALITYQuestionAnswering}.
We pick seven datasets that involve long texts, whose statistics are shown in \cref{tab:task}.
For more details about the data preprocessing, please refer to \cref{sec:app_data}.\footnote{Our codebase is available on \url{https://github.com/hiaoxui/long-range-transformers}.}

\begin{table}[t]
    \centering
    \begin{tabular}{lccc}
        \thick
        Dataset & Task & \#tokens & \#docs \\
         \hline
         Ontonotes & Coref. & 467 & 3493 \\
         TriviaQA & eQA & 2895 & 95k \\
         DocNLI & NLI & 399 & 1.44m \\
         SummFD & Summ. & 5.6k & 4.3k\\
         GovRep & Summ. & 7.9k & 19k\\
         Qasper & aQA & 3.7k & 5.7k\\
         QuALITY & aQA & 4.2k & 6.7k \\
         \thick
    \end{tabular}
    \caption{Task and dataset overview. The \#tokens is the number of tokens per doc on average.}
    \label{tab:task}
\end{table}

\section{Experiments}
\label{sec:exp}
\subsection{Coreference Resolution}
\label{sec:exp_coref}
Coreference resolution (coref.) is the task of identifying mention spans and clustering them into entities.
We consider multiple coreference strategies:  1) the widely used Coarse2Fine (C2F) method\footnote{We used the re-implementation by \citet{gardner2018AllenNLPDeepSemantic}.}~\citep{lee2018HigherorderCoreferenceResolution} which relies on span representations and 2) the current state-of-the-art method called Start2End (S2E)~\citep{kirstain2021CoreferenceResolutionSpan} that works on token representations.
The dataset we use is \textbf{Ontonotes 5.0}.
Transformers considered include \textbf{Longformer}, \textbf{XLNet}, and \textbf{Performer}.
We migrate the parameters of \textbf{SpanBERT} and \textbf{RoBERTa} to Performer and include results on these models as well.
We segment the input tokens into chunks with lengths of $L$
.
For models with global tokens, we lack a natural choice so we consider all tokens to be global.
For XLNet, we keep the memory of the same length of the segments (e.g. we keep a memory length of 256 for a model with segment length 256).\footnote{We adopt the same strategies with segmentation and memories in the remainder of the paper.}
The results are shown in \cref{tab:coref}.
Refer to \cref{sec:coref_full_exp} for complete results.

\begin{table}[t]
    \centering
    \begin{tabular}{l|l|ccc}
    \thick
    &Encoder & \lengths \\
    \hline
    \multirow{9}{*}{\rotatebox{90}{Model: Coarse2Fine}}&Longformer & 75.74 & 76.72 & 77.36 \\
    &Longformer$^G$ & 75.68 & 76.25 & 77.23 \\
    &BigBird & 75.95 & 76.78 & 77.64 \\
    &XLNet &74.57 & 74.48 & 74.33 \\
    &XLNet$^m$ & 74.73 & 75.76 & 76.29 \\
    &RoBERTa & 74.64 & 76.45 & 76.83  \\
    &RoBERTa$^p$ & 51.58 & 51.71 & 50.39  \\
    &SpanBERT & 75.04 & 75.84 & 76.59 \\
    &SpanBERT$^p$ & 52.46 & 52.06 & 50.51 \\
    \cline{2-5}
    &\longresult{Longformer}{76.77}{76.32}\\
    &\longresult{BigBird}{77.31}{77.57}\\
    \thick
    \multirow{9}{*}{\rotatebox{90}{Model: Start2End}}&Longformer & 74.77 & 76.27 & 77.73 \\
    &Longformer$^G$ & 74.15 & 76.19 & 77.41 \\
    &BigBird & 73.68 & 75.57 & 77.40\\
    &XLNet & 45.89 & 60.05 & 68.23 \\
    &XLNet$^m$ & 52.61 & 56.37 & 66.91 \\
    &RoBERTa & 71.96 & 76.27 & 77.78  \\
    &RoBERTa$^p$ & 40.06 & 42.35 & 41.69  \\
    &SpanBERT & 68.70 & 74.27 & 75.32 \\
    &SpanBERT$^p$ & 38.69 & 41.93 & 42.10 \\
    \cline{2-5}
    &\longresult{Longformer}{77.54}{77.57} \\
    &\longresult{BigBird}{77.43}{77.66}\\
    \thick 
    \end{tabular}
    \caption{
    Coref. experiment results. Numbers are averaged F1 (MUC, B$^3$, and CEAF$_{\phi_4}$).
    Longformer with $^G$ uses global tokens.
    XLNet with $^m$ uses recurrence memory.
    Encoders with $^p$ have their self-attention replaced with a Performer kernel.
    }
    \label{tab:coref}
\end{table}

Some observations are consistent across two coref. models. 
1) Though further pretrained upon RoBERTa, pattern-based methods do not show improvement over RoBERTa, even with longer attention range.
2) Models gain advantage when the segments get longer, but it is saturated when the segment length reaches 512. Distant contexts might not be exploited.
3) Performer-based models under-perform their corresponding non-kernelized models by a huge gap.
4) XLNet performs better with cached memory, but the performance gain is less observable when for shorter segments.
\footnote{The observations on XLNet and Performer are consistent across all the tasks in this paper.}

\begin{figure}[t]
    \centering
    \includegraphics[scale=0.64]{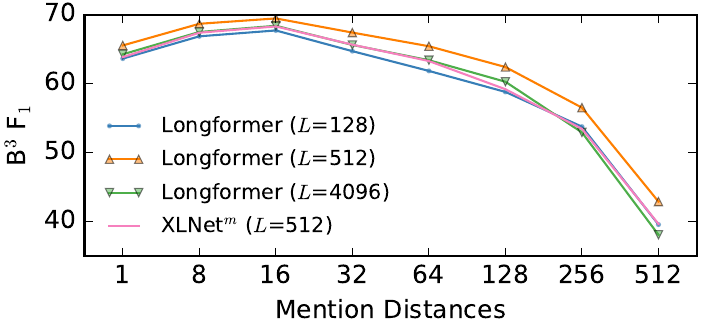}
    \caption{
    B$^3$ breakdown scores of 4 models for mention pairs with ranges from $[1, 8)$ to $[512, \infty)$.
    }
    \label{fig:b3}
\end{figure}

To further show the performance of those models on documents with different lengths, we conduct metric breakdown with a few typical configurations.
Instead of simply clustering the document according to the lengths, we propose a breakdown-version B$^3$ metric.
Given a mention distance range $[L_1, L_2]$, we calculate its corresponding B$^3$ value by only considering the mention pairs whose distances fall into this range.
The breakdown metrics are shown in \cref{fig:b3}.
We can see that the performance of all models follows the same trend and peaks at the $[16, 32)$ bucket.
Also, the graph shows that Longformer with longer context encoding does NOT benefit on distant mentions ($p$ value < 0.01). 
\footnote{We conduct significance test for the comparison between curves. Please refer to \cref{sec:app_sig} for more details.}
On the contrary, they suffer more from distance than shorter-context models, showing that long-range attention fails to capture long-distance information.

\subsection{Natural Language Inference}
\label{sec:exp_nli}
NLI is a classification task concerning a premise and hypothesis with variable lengths. The \textbf{DocNLI} dataset uses document-length inputs with binary labels (\texttt{entailment} and \texttt{not entailment}). 
We adopt the model proposed by \citet{yin2021DocNLILargescaleDataset},
and consider \textbf{Longformer}, \textbf{BigBird}, \textbf{Performer}, and \textbf{XLNet}.
We make the prediction on the \texttt{CLS} token, which is at the beginning for Longformer and the end for XLNet.
Since we are only interested in the encoding of \texttt{CLS},
we adopt the strategy of \citet{yin2021DocNLILargescaleDataset} which truncates the sequence to $L$ while preserving the hypothesis for Longformer and RoBERTa.
The results are shown in \cref{tab:nli}.

\begin{table}[t]
    \centering
    \begin{tabular}{l|ccccc}
         \thick
         Encoder & \lengths \\
         \thick
         XLNet & 29.95 & 40.39 & 24.31 \\
         XLNet$^m$ & 32.94 & 45.97 & 30.42 \\
         RoBERTa & 48.96 & 47.78 & 46.04 \\
         RoBERTa$^p$ & 17.83 & 24.91 & 23.65 \\
         Longformer & 29.11 & 25.73 & 45.28 \\
         BigBird & 28.95 & 24.71 & 31.72 \\
         \hline
         \longresult{Longformer}{45.96}{44.42} \\
         \longresult{BigBird}{33.58}{18.08} \\
         \thick
    \end{tabular}
    \caption{
    F1 scores on the test set for DocNLI.
    }
    \label{tab:nli}
\end{table}

\begin{figure}[t]
    \centering
    \includegraphics[scale=0.64]{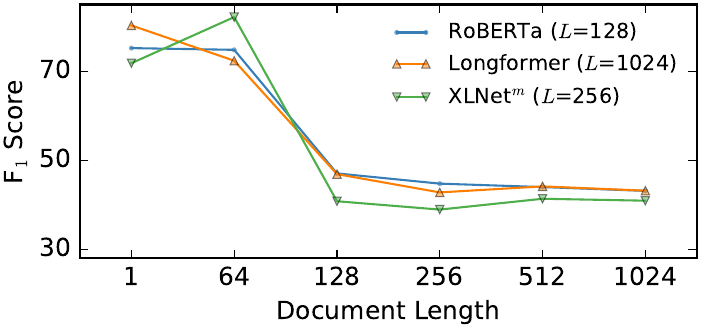}
    \caption{The breakdown analysis of DocNLI. We pick the best configuration for each model for brevity.}
    \label{fig:nli_bd}
\end{figure}

Observing \cref{tab:nli}, surprisingly, the best performance is achieved with short segments, though the optimal lengths vary from model to model.
Also, the performance of Longformer and BigBird is much lower than their baseline RoBERTa (c.f. \citet{yin2021DocNLILargescaleDataset}).
Observing the breakdown analysis in \cref{fig:nli_bd},
the performance of 3 best models follow the same trend w.r.t. the document length.\footnote{The graph looks less smooth than \cref{fig:b3}, possibly because DocNLI is made up of examples pulled from different datasets, which may have examples of different average lengths (cf. tab 1 in \citet{yin2021DocNLILargescaleDataset}).
Therefore the length of an example in DocNLI may correlate with different domains making up the dataset, which would interfere with our analysis.  
Future work will consider other datasets without this confounding concern.}
We speculate that all models are unable to comprehend the relationship between long documents,
and long-range attention does not bring any advantage.

\subsection{Question Answering}
\label{sec:exp_qa}

\begin{table}
\centering
\begin{tabular}{l|ccc}
\thick
Encoder  &  \lengths \\
\hline
Longformer & 54.26 & 58.83 & 63.88 \\
RoBERTa & 55.81 & 60.29 & 63.45 \\
RoBERTa$^p$ & 23.17 & 21.87 & 21.11 \\
BigBird & 55.28 & 59.39 & 63.51 \\
XLNet & 51.46 & 56.26 & 60.05 \\
XLNet$^m$ & 52.71 & 57.96 & 62.85 \\
\hline
\longresult{Longformer}{63.91}{63.66} \\
\longresult{Longformer$^G$}{\ \ \ -\ \ \ \ \ }{72.96} \\
\longresult{BigBird}{66.50}{71.78} \\
\thick
\end{tabular}
\caption{
Results for TriviaQA.
Longformer$^G$ indicates that the Longformer sets question as the global tokens. 
}
\label{tab:QA-1}
\end{table}

For question answering we experiment with  eQA (\textbf{TriviaQA}) and aQA  (\textbf{Qasper} and \textbf{QuALITY}).
\textbf{Performer}, \textbf{Longformer}, \textbf{BigBird}, and \textbf{XLNet} are tested for encoder-only tasks;
\textbf{BART} and Longformer-Encoder-Decoder (\textbf{LED}) is used for seq2seq tasks.
For Longformer$^G$ and BigBird, we set the question text (and candidate answers for QuALITY) as global tokens.\footnote{The global tokens of BigBird are fixed in the first 2 blocks, so we place the query at the beginning of the sequence.}
\textbf{TriviaQA} is a question answering dataset that involves extracting answer spans from reference documents. 
We adopt the method and codebase of \citet{joshi2017TriviaQALargeScale}. F1 is used as evaluation metrics.
\textbf{Qasper} addresses the QA task in the domain of academic papers and involves various answer types: extractive, abstractive, boolean, and unanswerable. 
We unify such tasks as an abstractive QA task~(cf. \citet{shaham2022SCROLLSStandardizedCompaRison}) and implement an LED-based decoder to generate answers.
F1 score is used for the evaluation of Qasper.
\textbf{QuALITY} is a multiple-choice QA task. 
Given a question and a passage, the task is to select the correct answer from several candidates. 
We regard it as a seq2seq problem, with the objective to predict the correct answer conditioned on the concatenation of  query, candidate answers, and passage.
During inference, the answer with the least perplexity is selected.
All results are shown in  \cref{tab:QA-1,tab:QA-2}, and the full results with more metrics are shown in \cref{sec:qa_full_exp}.

\begin{table}[t]
\centering
\begin{tabular}{l|l|cccc}
\thick
& Chunk & \color{blue}{512} & \color{blue}{1024} & \color{blue}{1536} & \color{blue}{$\infty$} \\
\hline
\multirow{3}{*}{\rotatebox{90}{Qasper}} & BART & 24.70 & 26.30 & - & - \\
& LED & 8.40 & 15.80 & 17.86 & 18.79 \\
& LED$^G$ & - & - & - & 28.64 \\
\hline
\multirow{3}{*}{\rotatebox{90}{QLTY}} & BART & 26.80 & 26.00 & - & - \\
& LED & 30.73 & 31.35 & 31.78 & 31.21 \\
& LED$^G$ & - & - & - & 29.87 \\
\thick
\end{tabular}
\caption{
Performance Qasper and QuALITY.  Qasper is evaluated with F1 and QuALITY with accuracy.
The results on BART are from \citet{shaham2022SCROLLSStandardizedCompaRison}.
}
\label{tab:QA-2}
\end{table}

Across all results, we find that larger receptive fields lead to better performance in most cases.
More importantly, setting queries as global tokens greatly benefits the performance on both eQA and aQA.
For QuALITY, it slightly hurt the performance to set both query and candidate as global tokens.
We reckon too many global tokens might introduce more noise, similar to the case of coref.
\begin{figure}[t]
    \centering
    \includegraphics[scale=0.64]{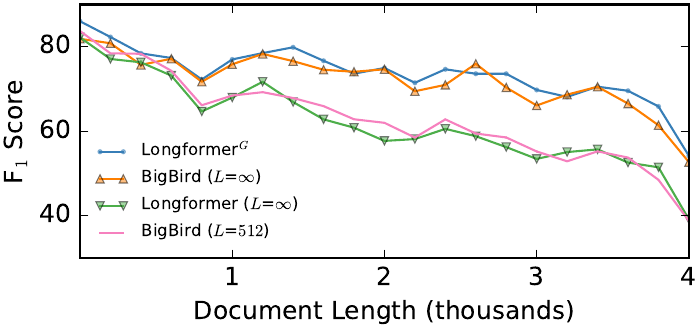}
    \caption{
    The performance of Longformer and BigBird on different lengths of TriviaQA documents.
    Note that Longformer$^G$ and BigBird ($L$=$\infty$) have global tokens.
    }
    \label{fig:extqa_bd}
\end{figure}

We speculate that the performance gain mostly comes from enhanced attention to the query,
which if further verified by the metric breakdown that is shown in \cref{fig:extqa_bd}. 
All models perform well on short texts, while models with global tokens obtain an observably greater advantage over the baseline models for longer documents ($p$ value < 0.01).
We think that the global token mechanism could help the model be less distracted on long texts via more attention on the queries~(\cref{sec:query}), which consequently improves the performance.

\subsection{Summarization}
\label{sec:exp_summ}
\begin{table}[t]
\centering
\begin{tabular}{l|l|cccc}
\thick
& Chunk & \color{blue}{512} & \color{blue}{1024} & \color{blue}{1536} & \color{blue}{$\infty$}\\
\hline
\multirow{2}{*}{\rotatebox{90}{SF}} & BART & 26.30 & 27.20 & - & - \\
& LED & 32.81 & 33.07 & 33.22 & 33.57 \\
\hline
\multirow{2}{*}{\rotatebox{90}{GR}} & BART & 45.60 & 47.90 & - & - \\
& LED & 53.86 & 54.13 & 54.83 & 56.60 \\
\thick
\end{tabular}
\caption{
Results for Summarization on SummFD and GovRep.
ROUGE unigram is used as the metric.
Results on BART are from \citet{shaham2022SCROLLSStandardizedCompaRison}.
}
\label{tab:summ}
\end{table}

As a typical seq2seq problem, 
we adopt \textbf{LED} to perform the summarization task.
We chunk the source sequence into segments (no segmentation for $L$ = $\infty$) to restrict the receptive range.
Intuitively, the summary may be benefited from the contextual representation with a broader view of the document.
Two datasets are used:
\textbf{SumScreen}  addresses the domain of TV shows. 
Following \citet{shaham2022SCROLLSStandardizedCompaRison}, we use the subset of ForeverDreaming (SummFD) consisting of 88 different shows.
The goal is to summarize the transcript of an episode, for which the recap is used as the ground truth summary.
\textbf{GovReport} is a long-document summarization dataset in the domain of government policies with human-written summaries. 
ROUGE \citep{lin2004ROUGEPackageAutomatic} is used as the evaluation metric.
Results are shown in \Cref{tab:summ}, and the full results are shown in \cref{sec:summ_full_exp}.

From \cref{tab:summ}, we observe slight superiority of the context-aware models.
Note that cross attention will attend to all source tokens whether we segment it or not,
but the intuition of summarization is to \emph{skim over} the document,
so we speculate that the performance improvement may be related to the \textit{selectivity} of the encoder-side attention, which is further analyzed in \cref{sec:seq2seq}. 
We do not conduct breakdown analysis for summarization because the sequence length directly contributes to the metric.

\section{Analysis}
\label{sec:analysis}

\subsection{The Attribution of Recurrence Memories}
\label{sec:attr}
Even if we know that distant contexts can help or hurt  performance,
it's still unclear \textit{how much} they contribute to the predictions.
One way to quantify this is \textit{attribution analysis}~\citep{simonyan2014DeepConvolutionalNetworks,li2017VisualizingUnderstandingNeural}.
Suppose $\ell$ is our loss function and $\mathbf{e}_i\in \mathbb{R}^d$ is the word embedding of the $i$-th token.
We use $\alpha_i$ to measure the attribution of the $i$-th token to the final prediction where
\begin{equation}
    \alpha_i = \left\| \frac{\partial \ell}{\partial \mathbf{e}_i} \right\|_l.
\end{equation}
We set $l=1$ to take the $L_1$ norm in practice, and the ground truth labels in the test set are used to calculate the gradient.
Intuitively, tokens with higher contribution have greater gradient norms.
Also, recurrent models like XLNet stop the gradient from being propagated back to the cached memory,
and we temporarily turn off this feature for analysis.

\begin{figure}[t]
    \centering
    \includegraphics[scale=0.64]{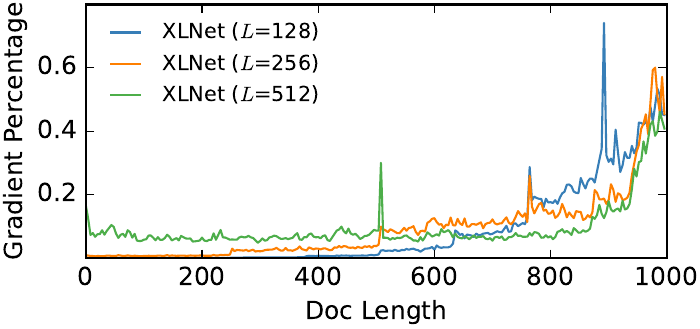}
    \caption{
    The gradient distribution over tokens for XLNet$^m$ on DocNLI documents.
    }
    \label{fig:nli_grad}
\end{figure}

We apply this method to XLNet on DocNLI,
where we pick 128 documents of lengths between 1,000 and 1,024.
We normalize the $\alpha_i$ over all tokens for each document, and take an average on each token index across all documents.
The results are shown in \cref{fig:nli_grad}.
Note that prediction is made in the last segment, and previous segments contribute only through the memory.
We see the attribution of tokens is stratified according to their segment lengths,
with a minor peak at the \texttt{CLS} token of each segment.
As we have more and more segments, the attribution of distant tokens to the final prediction becomes negligible.
For example, in the case of $L$=128, last segment made 53.4\% of attribution,
while the first segment made less than 0.01\%.

\subsection{Content Selection in Cross Attention}
\label{sec:seq2seq}
\begin{figure}[t]
    \centering
\includegraphics[scale=0.64]{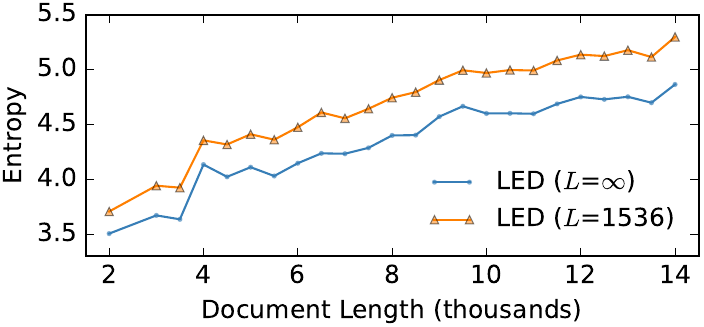}\\~\\
\includegraphics[scale=0.64]{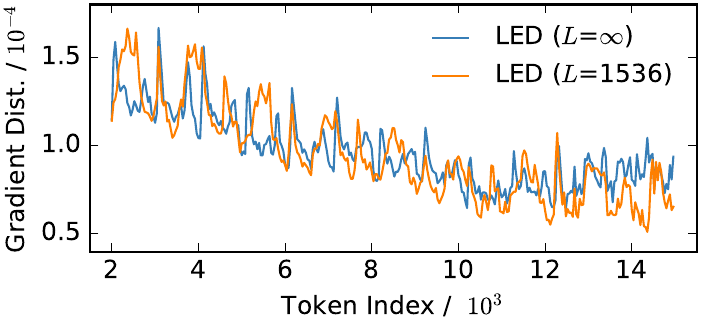}

    \caption{
    Attention distribution entropy (above) for documents with different lengths and attribution analysis (below) over source tokens for LED on SummFD.\label{fig:summ-fg}
    }
\end{figure}

The cross attention of LED attends to the whole document no matter if we segment the inputs or not,
thus we suppose source tokens should have similar attribution to decoding, which can be verified by \textit{attribution analysis} in \cref{fig:summ-fg}.
However, the crucial problem for summarization is whether the attention is \textit{selective},
given that only a portion of the document is helpful.
Therefore, we inspect the \textit{entropy}
\footnote{The distribution entropy is averaged over decoding tokens, attention heads, and transformer layers.}
of the cross attention distribution over the source tokens  \cref{fig:summ-fg}.

Reading the entropy curve, we find that the entropy of models without segmentation ($L$=$\infty$) is consistently lower ($p$ value < 0.01),
which can be translated to higher selectivity of cross attention and explains the superiority of LED ($L$=$\infty$) in \cref{tab:summ}.
Reading the gradient curve, we find that both models have a relatively uniform attribution over tokens with a slight slope on the left side due to the existence of short sequences.
This is reasonable because summarization requires the decoder to skim over the whole document, thus cross attention should not have a locality preference.

\subsection{Query-Guided Extraction and Decoding}
\label{sec:query}
\begin{figure}[t]
\includegraphics[scale=0.64]{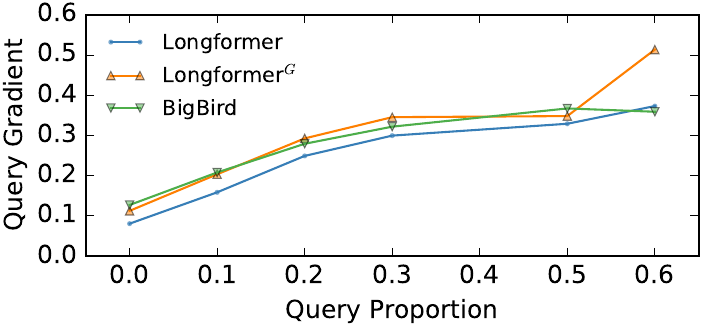}
\centering
\caption{
The accumulated gradient on query vs. the proportion of query, i.e. (query length / doc length) with Longformer and BigBird ($L$=$\infty$) on TriviaQA.
}
\label{fig:trivia-gradient}
\end{figure}

Different from other problems, the queries in QA can be treated as a ``guidance'' on how to read the documents.
To see if the encoders can exploit this structure,
we conduct attribution analysis on TriviaQA with Longformer and BigBird.
The results are shown on \cref{fig:trivia-gradient},
where we plot the relationship between the proportion of accumulated gradients on queries and the proportion of query tokens in the document.
It is clear that  models with global tokens pay more attention to queries (p value < $0.01$ except for one exception), 
which is consistent with the purpose of their design.

\begin{figure}[t]
    \centering
\includegraphics[scale=0.64]{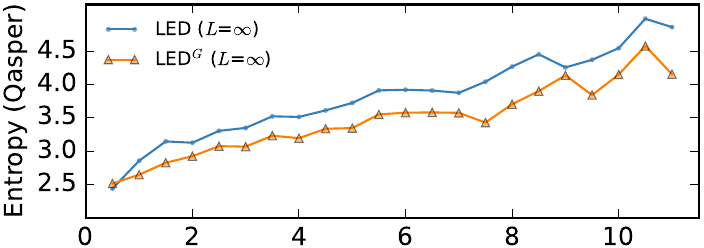}
\includegraphics[scale=0.64]{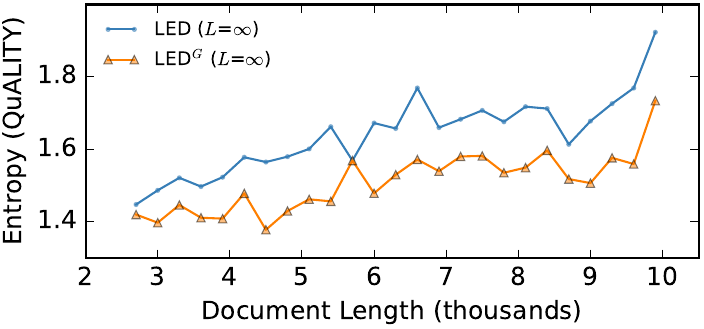}
    \caption{Entropy of source-side attention distribution vs. document length on Qasper and QuALITY.}
\label{fig:query_entropy}
\end{figure}

For scenarios where seq2seq decoding meets queries,
intuition suggests queries could instruct the cross-attention to attend to specific tokens, making the decoding more \textit{selective}.
From \cref{fig:query_entropy}, the entropy of attention distribution on source sequence against the doc length, we see
Longformer with global token unquestionably has lower entropy ($p$ value < 0.01), implying more targeted decoding.

\subsection{Error Accumulation of Kernel Methods}
\label{sec:kernel}
We find that Performer could not match the results of its prototypes (\cref{tab:coref,tab:nli,tab:QA-1}).
We suspect that the error incurred by the kernel approximation may not be acceptable for span-level tasks like coref.
We conducted another set of experiments: Instead of training the performer model from the checkpoints of SpanBERT,
we directly replace some layers from a fine-tuned SpanBERT model with Performer layers with their parameters preserved.
Experiments are conducted with C2F on Ontonotes 5.0.

\begin{figure}
    \centering
    \includegraphics[scale=0.64]{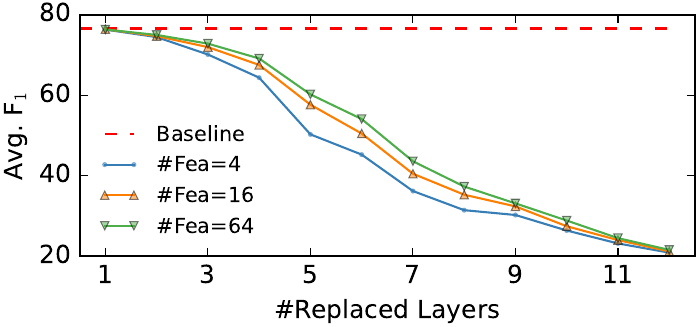}
    \caption{
    Results of SpanBERT ($L$=512) with layers replaced with Performer layers with ``\#Fea'' features.
    The baseline is the performance of the original model.
    }
    \label{fig:performer}
\end{figure}

In \cref{fig:performer}, we try to replace $P$ layers of SpanBERT in a top-down manner, where $P=1,2,\dots,12$.
We also try using different feature dimensions to exclude the possibility of insufficient features.
We find that although large dimension of random features brings marginal advantages to the performance,
Performer is not very sensitive to this factor.
Instead, the performance drops dramatically as we replace more layers,
from the baseline F1 78 to \textasciitilde 20 with all layers replaced. 
Based on our findings, we conclude that Performer is a good approximation for shallow transformers,
even with very low-dimensional random features.
However, as we stack more transformer layers, the accumulated errors can be unacceptable,
which leads to a failure in the performance.

\section{Experiment Confounders}
\label{sec:confounder}
\begin{figure}
    \centering
    \includegraphics[scale=0.64]{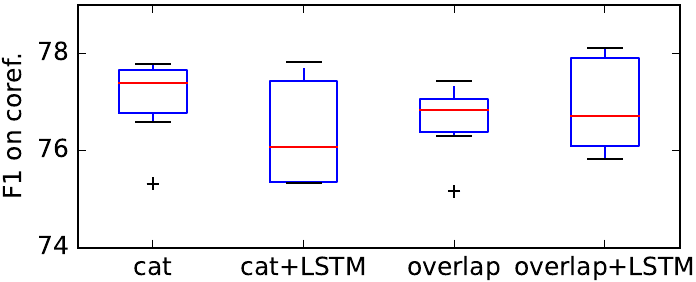}
    \caption{
    Results of different pooling strategies.
    }
    \label{fig:pooling}
\end{figure}

\paragraph{Pretraining and Adaptation}
The purpose of this paper is to evaluate the effectiveness of different long-range attention approaches by reducing the confounder of pre-training,
but it is still unclear how the results would change if we pre-train those models with the same configuration from scratch.
Unfortunately, it requires much more resources and introduces more confounders of training settings,
and we adopt the most straightforward way to ablate the long-range attention or migrate parameters.
For each model-task pair, we make the most natural choice, e.g. setting queries as global tokens, trying to minimize human biases of model adaptations.

\paragraph{Pooling Strategies} 
Concatenating the representation of segments, while being natural and commonly used, is not the only pooling strategy.
As a comparison, we incorporate the results of other 3 solutions:
1) Split the documents into segments of $L$ tokens with $L/2$ overlapped between adjacent segments~\citep{joshi2019BERTCoreferenceResolution};
2) Stack an LSTM layer over the Transformer representation of the segment ;
3) A combination of above methods.
We conduct experiments on the coreference resolution task with both C2F and S2E solution and many variants of transformers as our encoder. 
We leave the experiment details and discussions in \cref{sec:app_pooling} and show a brief result in \cref{fig:pooling}.
We have similar observation with \citet{joshi2019BERTCoreferenceResolution} that overlapping does not bring performance improvement.
Moreover, though introducing more parameters,
a stacked LSTM even hurts the performance.
We conclude that pooling strategies do not affect our analysis in \cref{sec:analysis}.

\section{Discussion}
Researchers have proposed many innovative methods for efficient self-attention over long sequences. The key ideas work as desired in certain cases, though we demonstrate several drawbacks.

Surprisingly, \textbf{pattern-based methods}, as the most popular approach, are not necessarily benefited from long-range attention in the general case.
Larger sliding windows are helpful, but the benefit would quickly saturate or become negative~(\cref{tab:coref}).
However, when a small portion of \emph{guidance text} (e.g. query in QA) exists, setting it as \emph{global tokens} can make it more attended and significantly boost the performance~(\cref{sec:query}).
When such text doesn't exist, setting all tokens as global would hurt the performance~(\cref{tab:coref}).
Moreover, we find that long-range attention and global tokens are correlated with the \emph{selectivity} of \emph{seq2seq} problems,
which consequently helps the decoding.

The memory of \textbf{recurrence models} generally improves performance,
proving historical hidden states are beneficial for transformers in various tasks.
However, XLNet does not fully exploit the history tokens,
with distant information less attended~(\cref{sec:attr}).
We speculate that it is because XLNet is pretrained to predict masked tokens, which does not frequently need the participation of long-range context~\citep{sun2021LongRangeLanguageModels}.
Also, the stop-gradient trick may hinder the model from effectively attending to memories.

The approximation of \textbf{kernel-based methods} works very well for shallow networks, but faces serious error accumulation problems when transformer layers are deeply stacked, which cannot be remedied by having high-dimensional random features~(\cref{sec:kernel}).
The resulting performance drop is not acceptable even for the ``base'' version of transformer encoders with 12 layers~(\cref{tab:coref}).

\section{Conclusion}
We conduct experiments with various long-range transformers on  NLP tasks that involve long sequences,
trying to fairly evaluate their long-range attention ability.
While some methods are validated on certain tasks,
we also find some previously unrecognized drawbacks.
We further analyze the attention behaviors of these transformers with metric breakdown, attribution analysis, and entropy analysis,
revealing the performance of those models might be correlated with the attribution of distant tokens, selectivity of attentions, or the approximation errors.
We hope our work would shed light on the future development of long-range transformers.

\paragraph{Model Selection Takeaways}
Based on our findings, we have the following suggestions.
For common tasks, such as sequence classification or token-level prediction, it is still competitive to chunk the inputs and apply short-range transformers.
When explicit guiding text, such as queries, exists, pattern-based models with global token mechanism is preferred.
For seq2seq problems, long-range transformers with pretrained checkpoints deliver better performance.

\section*{Acknowledgement}

We appreciate the proofreading done by Patrick Xia, Marc Marone, Nils Holzenberger, Elias Stengel-Eskin, Yunmo Chen, and Zhengping Jiang
. Thanks to the anonymous reviewers for their valuable feedback.

This work was supported in part by IARPA BETTER (\#2019-19051600005). The views and conclusions contained in this work are those of the authors and should not be interpreted as necessarily representing the official policies, either expressed or implied, or endorsements of ODNI, IARPA, or the U.S. Government. The U.S. Government is authorized to reproduce and distribute reprints for governmental purposes.

\section{Limitations}

\paragraph{Energy Cost}

Our experiments involve a massive amount of training with many transformers on various tasks.
Although we don't conduct any pretraining, the energy cost is still non-negligible.
However, our hope is our findings here would enable others to more efficiently select a particular architecture for their task, rather than reproducing the work done here.

\paragraph{Experimental Bias}
Due to the lack of pretrained checkpoints for general purposes,
we focus on representative instead of each type of transformer variant.
It is possible that these observations are particular to specific artifacts and implementations considered here.
Our goal is foremost to provide a roadmap for continued study on questions raised in this article,
with new architectures being evaluated in the future by model developers themselves.
For similar reasons, existing artifacts are biased towards English, 
as are many of the datasets employed in this study. 
We do not believe our findings are specific to English, but it remains for future work in long range transformer evaluation to extend our analysis into multilingual conditions.

\paragraph{Language Bias}
For similar reasons, existing artifacts are biased towards English, 
as are many of the datasets employed in this study. 
We do not believe our findings are specific to English, but it remains for future work in long-range transformer evaluation to extend our analysis into multilingual conditions.

\bibliography{ref}  
\bibliographystyle{acl_natbib}

\clearpage
\appendix

\section{Data Preprocessing}
\label{sec:app_data}
The coref task usually consumes a large amount of GPU memory.
For the experiments on Ontonotes, we truncate the sequence longer than 2000 tokens (for c2f model) or 1400 tokens (for s2e model) for the memory concern.
The truncation was applied only to the training set.

We didn't do any pre-processing steps for DocNLI~\citep{yin2021DocNLILargescaleDataset},
except for the truncation that we discussed in \cref{sec:exp}.
Notably, DocNLI is an aggregated dataset constructed from ANLI~\citep{nie2020AdversarialNLINew}, SQuAD~\citep{rajpurkar2016SQuAD100000}, DUC2001, CNN/DailyMail~\citep{nallapati2016AbstractiveTextSummarization}, and Curation~\citep{curation2020CurationCorpusBase}.
Documents from different sources may be distinguishable from their lengths (refer to tab 1 in \citet{yin2021DocNLILargescaleDataset}.

For TriviaQA, we use the scripts of Longformer~\cite{beltagy2020LongformerLongDocumentTransformer} to pre-process the data.
There is no further modifications on the data.

We adopt the dataset of QuALITY and GovReport from \citet{shaham2022SCROLLSStandardizedCompaRison}, which picked long sequences from those datasets.

For Qasper and SummScreen, we simply adopt the original dataset and scripts for preprocessing.

\section{Implementation Details of Transformers}
\label{sec:app_encoder}
In this section, we discuss the details of modification we did to the transformers.

\paragraph{Recurrence-Based Methods}
We tested XLNet for recurrence-based methods.
We adopt the codebase of Huggingface as the base model,
and followed the common strategy to stop gradient from being propagated back into the cached memory.
For each segment of the recurrence, 
we appended two special tokens \texttt{SEP} and \texttt{CLS} to the sequence,
making it like an ordinary input sequence to the XLNet model
except for the possible existence of memory states.
We concatenated the token representations after recurrences and remove the special tokens from all but the last segment.
Empirically, we found that having special tokens can significantly boost the performance.

\paragraph{Pattern-Based Methods}
For the pattern-based methods, 
when we chunked the input sequence into segments,
we appended the \texttt{SEP} as we did for the XLNet, and prepended the \texttt{CLS} token as a convention of other transformers.
After concatenation, we removed all the special tokens except for the first \texttt{CLS} the last \texttt{SEP},
which made it structurally similar to the outputs of non-segmented transformers.
What's more, for sliding window mechanism, we might reduce the window size to segment length if needed to save compute and memory.

\paragraph{Kernel-Based Methods}
We tested Performer~\citep{choromanski2021RethinkingAttentionPerformers} as a representative of kernel-based methods.
Given that training from random initialization would lead to suboptimal results, we migrate the parameters from base models (e.g. BERT, RoBERTa) to kernel methods.
In detail, we replace the self-attention layers of the base models with kernels.
Given that Performer does not require any additional parameters, except for the orthogonal random feature vectors in the FAVOR+ mechanism.
Following the default implementation of the fast transformer codebase\footnote{\url{https://github.com/idiap/fast-transformers}},
we set the feature dimension as the query dimension by default for main experiments,
though we found the performance isn't sensitive to those features in \cref{sec:kernel}.

\section{Experiments Details}
\label{sec:app_exp}
In this section, we introduce the details of our experiments in the main paper, including hyper-parameters, training strategies, data split and loading, and other configurations that are necessary to replicate our results.

\paragraph{Computational Resources}
All of our experiments were done with NVIDIA RTX 6000 GPU with 24GB of memory.
We did most of the experiments with single cards, except for TriviaQA, for which we used multi-GPU training with 4 cards.

\paragraph{Coreference Resolution}
We used two models: coarse2fine (C2F)~\citep{lee2018HigherorderCoreferenceResolution} and start2end (S2E)~\citep{kirstain2021CoreferenceResolutionSpan}.
For C2F model, we used the codebase reimplemented by AllenNLP~\citep{gardner2018AllenNLPDeepSemantic} for its flexibility on encoder exchange.
We used the official codebase of S2E model for other experiments.
We didn't change any hypermeters except for the difference on the encoders.
We adopted the same training strategy of these repos without any modifications,
i.e. we train the model with certain epochs (40 for C2F and 129 for S2E) or until convergence and pick the model performed best on the dev set for evaluation.
The typical training time of coref models was 10h for C2F and 24h for S2E.

\paragraph{Natural Language Inference}
Due to the size of DocNLI dataset (942k training and 234k dev examples), it's infeasible to adopt the common training strategies.
Instead, we train the model with mini-batch gradient descent with a batch size of 4 for only one epoch.
Because the training and dev set are too large to fit into CPU memory, 
we split the training set into smaller datasets consisting of 32768 examples, and iterate over training and dev set during training.
We pick the model with best performance on the dev set (not the whole set but one iteration) for test.
We use the whole test set consisting of 267k examples for the final evaluation.
We adopt the same architecture as the model used in \citet{yin2021DocNLILargescaleDataset} and reimplement it with AllenNLP~\citet{gardner2018AllenNLPDeepSemantic}.
The typical training time is around 2 days and the test time is around 4 hours.

\paragraph{Question Answering}
For TriviaQA, we adopted the training scripts and hyperparameters used by \citet{beltagy2020LongformerLongDocumentTransformer} except for that we set the training batch size as 4 and number of epochs as 8. 
The performance is evaluated on the dev set and we pick the best checkpoint with patience of 3.

For Qasper, we follow the training scripts and hyper-parameters in its official repository.
\footnote{\url{https://github.com/allenai/qasper-led-baseline}}
We disable the evidence setting, and extend the training to a maximum epoch of 20.
The performance is evaluated on the dev set with patience of 5.

For QuALITY, we adopt the LED model with ``allenai/led-base-16384'' configuration from Huggingface.
\footnote{\url{https://huggingface.co/allenai/led-base-16384}}
We concatenate the question, candidate answers, and passages as the encoder input, and the correct answer as the decoder input for training.
During inference, we feed each candidate answer as the decoder input, and consider the one with the lowest perplexity as the predicted answer.
We use a warmup steps of 1000 and learning rate of $5\times 10^{-5}$ for training. 
Evaluation is performed on the dev set with a patience of 5.

\paragraph{Summarization}
For both the SummScreen and GovReport datasets, we use the LED model with ``allenai/led-base-16384'' configuration. 
We use a warmup steps of 1000 and learning rate of $5\times 10^{-5}$ for training and patience of 5 for testing.
GovReport is evaluated on the dev set and SummScreen is evaluated on its official test set.

\section{Significance Test}
\label{sec:app_sig}
For the comparison between curves in the \cref{sec:analysis}, we conduct significance test using bootstrapping methods to verify our conclusions.
Let $D$ be the test set for a task.
For the performance comparison between two configurations, we sample a new $D^*$ from $D$ with replacement and we keep $|D^*|=|D|$.
We treat the event ``configuration A performs better than B'' as a Bernoulli random variable $P$,
and compute the probability of the null hypothesis $P<0.5$ as the $p$ value.
We sample $B=1024$ test sets for each comparison.
If multiple significance tests are conducted, we only report the larges value that we obtain.
\footnote{
For the curves in \cref{fig:trivia-gradient}, we exclude one exception case at $x=0.6$.
For the curves in \cref{fig:query_entropy}, we exclude one exception case at $x=500$.
}

For example, in \cref{fig:b3}, we claim that the performance of Longformer ($L$=512) is better than any other encoders regardless of the mention distances.
To verify it, we conduct significance test between Longformer ($L$=512) and other 3 encoders for every mention distance.
The greatest $p$ value among 24 $p$ tests is smaller than 0.01, so our claim is secured by our significance test.

\section{More Pooling Strategies}
\label{sec:app_pooling}

We conduct experiments with 4 transformers, including 2 short-range transformers (RoBERTa and SpanBERT) and 2 long-range transformers (Longformer and BigBird) on the coreference resolution task.
We set $L$=512, which is the maximum acceptable length for short-range transformers.

The full results are shown in \cref{tab:detail_pooling},
and a box plot can be found in \cref{fig:pooling}.
In overall,
we have similar observations as \citet{joshi2019BERTCoreferenceResolution} that overlapped segments do not offer improvements on the performance.
Similar findings can be found for LSTM settings and the combination of them.
More importantly, the performance difference of \cref{tab:detail_pooling} is consistently with \cref{tab:detail_coref_c2f,tab:detail_coref_s2e} except for a few outliers.
Thus, we conclude that direct concatenation is already enough to exploit the pretrained transformers,
and changing pooling strategies do not greatly interfere our analysis.

\begin{table*}[t]
    \centering
    \begin{tabular}{@{\extracolsep{2pt}}l|lccccccccccc@{}}
    \thick
    &\multirow{2}{*}{Encoder} & \multicolumn{3}{c}{MUC} & \multicolumn{3}{c}{B$^3$} & \multicolumn{3}{c}{CEAF$_{\phi_4}$} & \multirow{2}{*}{Avg.}\\
    \cline{3-5} \cline{6-8} \cline{9-11}
    && P & R & F1 & P & R & F1 & P & R & F1 & \\
\hline
\multirow{8}{*}{\rotatebox{90}{Model: Coarse2Fine}} & RoBERTa & 81.6&85.0&83.3&72.1&78.2&75.1&72.1&72.2&72.2&76.8 \\
& RoBERTa$_\textrm{\small overlap}$ & 82.9&84.5&83.7&73.1&77.1&75.0&73.5&71.7&72.5&77.1 \\
& RoBERTa$^\textrm{\small LSTM}$ & 84.3&83.9&84.1&75.0&76.2&75.6&74.5&71.0&72.7&77.5 \\
& RoBERTa$_\textrm{\small overlap}^\textrm{\small LSTM}$ & 84.1&84.8&84.5&75.4&77.4&76.4&74.5&72.4&73.5&78.1 \\
& Longformer & 82.4&85.3&83.8&73.0&78.6&75.7&72.6&72.5&72.5&77.4 \\
& Longformer$_\textrm{\small overlap}$ & 82.4&83.8&83.1&72.9&76.0&74.4&72.3&70.5&71.4&76.3 \\
& Longformer$^\textrm{\small LSTM}$ & 84.2&83.8&84.0&75.3&75.8&75.5&73.8&71.7&72.8&77.4 \\
& Longformer$_\textrm{\small overlap}^\textrm{\small LSTM}$ & 84.3&84.2&84.2&75.4&76.6&76.0&74.4&72.3&73.4&77.9 \\
& BigBird & 81.5&86.9&84.1&71.5&80.9&75.9&72.7&73.1&72.9&77.6 \\
& BigBird$_\textrm{\small overlap}$ & 83.0&84.3&83.6&73.9&76.7&75.3&72.4&72.0&72.2&77.0 \\
& BigBird$^\textrm{\small LSTM}$ & 84.2&84.4&84.3&75.0&77.1&76.0&74.0&72.1&73.1&77.8 \\
& BigBird$_\textrm{\small overlap}^\textrm{\small LSTM}$ & 84.5&84.3&84.4&75.7&76.8&76.2&74.8&72.2&73.5&78.0 \\
& SpanBERT & 83.3&82.9&83.1&74.4&74.8&74.6&72.6&71.7&72.1&76.6 \\
& SpanBERT$_\textrm{\small overlap}$ & 83.1&83.0&83.0&74.4&75.0&74.7&72.2&72.2&72.2&76.7 \\
& SpanBERT$^\textrm{\small LSTM}$ & 83.4&82.8&83.1&74.3&74.7&74.5&72.8&70.9&71.8&76.5 \\
& SpanBERT$_\textrm{\small overlap}^\textrm{\small LSTM}$ & 83.6&82.6&83.1&74.4&74.6&74.5&72.6&70.8&71.7&76.4 \\
\hline
\multirow{8}{*}{\rotatebox{90}{Model: Start2End}} & RoBERTa & 85.7&82.6&84.1&78.1&74.3&76.2&75.2&70.9&73.0&77.8 \\
& RoBERTa$_\textrm{\small overlap}$ & 85.0&82.6&83.8&77.7&73.5&75.6&74.4&71.5&73.0&77.4 \\
& RoBERTa$^\textrm{\small LSTM}$ & 83.8&81.4&82.6&75.1&72.3&73.7&72.1&69.4&70.8&75.7 \\
& RoBERTa$_\textrm{\small overlap}^\textrm{\small LSTM}$ & 83.7&82.9&83.3&75.6&74.6&75.1&73.7&71.4&72.5&77.0 \\
& Longformer & 85.5&82.6&84.0&78.0&74.5&76.2&75.2&70.7&72.9&77.7 \\
& Longformer$_\textrm{\small overlap}$ & 83.9&82.2&83.1&75.5&73.3&74.4&73.8&69.9&71.8&76.4 \\
& Longformer$^\textrm{\small LSTM}$ & 84.0&80.8&82.3&75.7&71.2&73.4&72.5&68.3&70.3&75.3 \\
& Longformer$_\textrm{\small overlap}^\textrm{\small LSTM}$ & 83.7&81.6&82.6&75.3&72.5&73.8&72.5&69.5&71.0&75.8 \\
& BigBird & 85.7&82.3&84.0&77.9&73.9&75.9&75.7&69.4&72.4&77.4 \\
& BigBird$_\textrm{\small overlap}$ & 84.1&82.4&83.2&76.3&74.2&75.2&74.7&70.8&72.7&77.1 \\
& BigBird$^\textrm{\small LSTM}$ & 83.2&81.7&82.4&74.2&72.5&73.3&72.1&68.5&70.3&75.3 \\
& BigBird$_\textrm{\small overlap}^\textrm{\small LSTM}$ & 83.7&81.9&82.8&75.5&73.0&74.2&73.3&69.9&71.6&76.2 \\
& SpanBERT & 83.5&81.3&82.4&74.6&71.9&73.2&72.3&68.5&70.3&75.3 \\
& SpanBERT$_\textrm{\small overlap}$ & 82.9&81.2&82.0&74.0&72.3&73.2&72.2&68.5&70.3&75.2 \\
& SpanBERT$^\textrm{\small LSTM}$ & 70.4&58.9&64.1&47.8&44.7&46.2&59.6&26.7&36.9&49.1 \\
& SpanBERT$_\textrm{\small overlap}^\textrm{\small LSTM}$ & 68.3&61.5&64.7&43.6&47.8&45.6&59.3&26.3&36.5&48.9 \\
\thick
    \end{tabular}
    \caption{The full results of all experiments with different pooling strategies. All models use the segment length $L$=512. Models with superscript ``LSTM'' indicate it uses LSTM, and subscript ``overlap'' indicates it uses overlapped concatenation method. Note that both methods can be applied in the meantime.}
    \label{tab:detail_pooling}
\end{table*}

\section{Full Experiment Results}
\label{sec:app_full_exp}
In this section, we list the full results of all the experiments in the main paper.

\subsection{Coreference Resolution}
\label{sec:coref_full_exp}
The full results of coreference resolution are shown in \cref{tab:detail_coref_c2f,tab:detail_coref_s2e}.
We use MUC~\citep{vilain1995ModeltheoreticCoreferenceScoring}, B$^3$~\citep{bagga1998AlgorithmsScoringCoreference}, and CEAF$_{\phi_4}$~\citep{luo2005CoreferenceResolutionPerformance} as the evaluation metrics.
Following the convention, we use the ``Avg.'' as the main metric, which is an average among the F1 score of 3 metrics.
All the results are reported on the test set.

\begin{table*}[t]
    \centering
    \begin{tabular}{@{\extracolsep{2pt}}lccccccccccc@{}}
    \thick
    \multirow{2}{*}{Encoder} & \multicolumn{3}{c}{MUC} & \multicolumn{3}{c}{B$^3$} & \multicolumn{3}{c}{CEAF$_{\phi_4}$} & \multirow{2}{*}{Avg.}\\
    \cline{2-4} \cline{5-7} \cline{8-10}
    & P & R & F1 & P & R & F1 & P & R & F1 & \\
    \hline
BigBird~{\small($L$=128)} & 80.5 & 85.5 & 83.0 & 69.8 & 78.5 & 73.9 & 70.8 & 71.2 & 71.0 & 75.9 \\
BigBird~{\small($L$=256)} & 81.1 & 86.2 & 83.6 & 70.5 & 79.8 & 74.9 & 72.1 & 71.7 & 71.9 & 76.8 \\
BigBird~{\small($L$=512)} & 81.5 & 86.9 & 84.1 & 71.5 & 80.9 & 75.9 & 72.7 & 73.1 & 72.9 & 77.6 \\
BigBird~{\small($L$=1024)} & 82.2 & 85.5 & 83.8 & 72.8 & 78.4 & 75.5 & 72.7 & 72.5 & 72.6 & 77.3 \\
BigBird~{\small($L$=4096)} & 81.8 & 87.0 & 84.3 & 71.5 & 81.0 & 76.0 & 72.7 & 73.2 & 73.0 & 77.7 \\
Longformer~{\small($L$=128)} & 81.7 & 83.7 & 82.7 & 71.6 & 75.8 & 73.7 & 71.3 & 70.4 & 70.9 & 75.7 \\
Longformer$^G$~{\small($L$=128)} & 81.1 & 84.3 & 82.7 & 70.6 & 76.5 & 73.4 & 71.0 & 70.9 & 70.9 & 75.7 \\
Longformer~{\small($L$=256)} & 81.6 & 85.2 & 83.4 & 71.8 & 78.1 & 74.8 & 71.6 & 72.3 & 72.0 & 76.7 \\
Longformer$^G$~{\small($L$=256)} & 81.4 & 84.8 & 83.1 & 71.4 & 77.6 & 74.4 & 71.0 & 71.7 & 71.3 & 76.3 \\
Longformer~{\small($L$=512)} & 82.4 & 85.3 & 83.8 & 73.0 & 78.6 & 75.7 & 72.6 & 72.5 & 72.5 & 77.4 \\
Longformer$^G$~{\small($L$=512)} & 82.6 & 85.0 & 83.8 & 73.2 & 77.9 & 75.5 & 72.4 & 72.4 & 72.4 & 77.2 \\
Longformer~{\small($L$=1024)} & 82.1 & 84.9 & 83.5 & 72.3 & 77.7 & 74.9 & 72.0 & 72.0 & 72.0 & 76.8 \\
Longformer~{\small($L$=4096)} & 82.0 & 84.2 & 83.1 & 72.4 & 76.2 & 74.3 & 71.6 & 71.6 & 71.6 & 76.3 \\
RoBERTa~{\small($L$=128)} & 81.4 & 82.5 & 81.9 & 70.7 & 73.9 & 72.2 & 71.4 & 68.2 & 69.7 & 74.6 \\
RoBERTa$^p$~{\small($L$=128)} & 69.4 & 57.7 & 63.0 & 55.9 & 42.5 & 48.3 & 49.9 & 38.4 & 43.4 & 51.6 \\
RoBERTa~{\small($L$=256)} & 82.0 & 84.5 & 83.2 & 72.2 & 77.2 & 74.6 & 72.3 & 70.7 & 71.5 & 76.5 \\
RoBERTa$^p$~{\small($L$=256)} & 68.7 & 57.9 & 62.9 & 55.7 & 43.1 & 48.6 & 49.3 & 39.2 & 43.7 & 51.7 \\
RoBERTa~{\small($L$=512)} & 81.6 & 85.0 & 83.3 & 72.1 & 78.2 & 75.1 & 72.1 & 72.2 & 72.2 & 76.8 \\
RoBERTa$^p$~{\small($L$=512)} & 68.0 & 56.1 & 61.5 & 55.1 & 40.8 & 46.9 & 48.4 & 38.4 & 42.8 & 50.4 \\
SpanBERT~{\small($L$=128)} & 82.0 & 82.2 & 82.1 & 72.0 & 73.7 & 72.8 & 71.5 & 69.0 & 70.2 & 75.0 \\
SpanBERT$^p$~{\small($L$=128)} & 70.6 & 56.8 & 63.0 & 58.1 & 42.9 & 49.4 & 50.8 & 40.5 & 45.1 & 52.5 \\
SpanBERT~{\small($L$=256)} & 82.7 & 82.8 & 82.7 & 73.0 & 74.1 & 73.5 & 71.9 & 70.6 & 71.3 & 75.8 \\
SpanBERT$^p$~{\small($L$=256)} & 70.0 & 56.4 & 62.5 & 58.5 & 41.7 & 48.7 & 50.1 & 40.8 & 45.0 & 52.1 \\
SpanBERT~{\small($L$=512)} & 83.3 & 82.9 & 83.1 & 74.4 & 74.8 & 74.6 & 72.6 & 71.7 & 72.1 & 76.6 \\
SpanBERT$^p$~{\small($L$=512)} & 67.6 & 55.7 & 61.1 & 56.2 & 40.8 & 47.3 & 47.2 & 39.8 & 43.2 & 50.5 \\
XLNet~{\small ($L$=128, $m$=0)} & 81.6 & 82.7 & 82.1 & 71.2 & 73.4 & 72.3 & 70.0 & 68.6 & 69.3 & 74.6 \\
XLNet~{\small ($L$=128, $m$=128)} & 81.7 & 82.6 & 82.1 & 71.9 & 73.3 & 72.6 & 69.9 & 69.0 & 69.5 & 74.7 \\
XLNet~{\small ($L$=256, $m$=0)} & 79.4 & 84.2 & 81.7 & 68.5 & 76.0 & 72.0 & 68.5 & 70.9 & 69.7 & 74.5 \\
XLNet~{\small ($L$=256, $m$=256)} & 84.3 & 81.8 & 83.0 & 75.4 & 72.2 & 73.8 & 72.2 & 68.9 & 70.5 & 75.8 \\
XLNet~{\small ($L$=512, $m$=0)} & 79.0 & 85.2 & 82.0 & 67.4 & 77.4 & 72.1 & 69.1 & 68.8 & 69.0 & 74.3 \\
XLNet~{\small ($L$=512, $m$=512)} & 82.1 & 84.1 & 83.1 & 72.5 & 76.1 & 74.3 & 72.8 & 70.3 & 71.5 & 76.3 \\
\thick
    \end{tabular}
    \caption{Full results on Ontonotes with the coarse2fine model. $L$ is the segment length used to chunk the text. $m$ is the memory length used for the XLNet model. $^G$ denotes that the global tokens are used. $^p$ denotes that th self-attention computation is replaced with Performer kernels.}
    \label{tab:detail_coref_c2f}
\end{table*}

\subsection{Question Answering}
\label{sec:qa_full_exp}

The full experiment results on TriviaQA is shown in \cref{tab:detail_eqa}.
We use both F1 and exact match (EM) as the metrics.
A few cells are left blank because of the constraints of the transformers.

\begin{table*}[t]
    \centering
    \begin{tabular}{@{\extracolsep{2pt}}lccccccccccc@{}}
    \thick
    \multirow{2}{*}{Encoder} & \multicolumn{2}{c}{$L$=128} & \multicolumn{2}{c}{$L$=256} & \multicolumn{2}{c}{$L$=512} & \multicolumn{2}{c}{$L$=1024} & \multicolumn{2}{c}{$L$=$\infty$}\\
    \cline{2-3} \cline{4-5} \cline{6-7} \cline{8-9} \cline{10-11}
    & F1 & EM & F1 & EM & F1 & EM & F1 & EM & F1 & EM  \\
    \hline
Longformer & 54.26&50.02&58.83&54.48&63.88&59.13&63.91&58.91&63.41&58.89 \\
Longformer$^G$ & \ \ \ -\ \ \ \ \ &\ \ \ -\ \ \ \ \ &\ \ \ -\ \ \ \ \ &\ \ \ -\ \ \ \ \ &\ \ \ -\ \ \ \ \ &\ \ \ -\ \ \ \ \ &\ \ \ -\ \ \ \ \ &\ \ \ -\ \ \ \ \ &72.96&67.88 \\
RoBERTa & 55.81&50.73&60.29&56.11&63.45&58.84&\ \ \ -\ \ \ \ \ &\ \ \ -\ \ \ \ \ &\ \ \ -\ \ \ \ \ &\ \ \ -\ \ \ \ \  \\
RoBERTa$^p$ & 23.17&16.80&21.87&15.56&21.11&15.09&\ \ \ -\ \ \ \ \ &\ \ \ -\ \ \ \ \ &\ \ \ -\ \ \ \ \ &\ \ \ -\ \ \ \ \  \\
BigBird & 55.28&50.66&59.39&54.34&63.51&58.50&66.50&61.15&71.78&66.86 \\
XLNet & 51.46&47.10&56.26&52.08&60.05&55.62&\ \ \ -\ \ \ \ \ &\ \ \ -\ \ \ \ \ &\ \ \ -\ \ \ \ \ &\ \ \ -\ \ \ \ \  \\
XLNet$^m$ & 52.71&48.03&57.96&52.93&62.85&58.13&\ \ \ -\ \ \ \ \ &\ \ \ -\ \ \ \ \ &\ \ \ -\ \ \ \ \ &\ \ \ -\ \ \ \ \  \\

\thick
    \end{tabular}
    \caption{Full results on TriviaQA. We adopt the same notation as we used in \cref{tab:detail_coref_c2f}.}
    \label{tab:detail_eqa}
\end{table*}

\subsection{Summarization}
\label{sec:summ_full_exp}
The full results on summarization is shown in \cref{tab:full_summ}. We used ROUGE~\citep{lin2004ROUGEPackageAutomatic} as the metric. R1, R2, and R3 stands for ROUGE unigram, bigram, and longest common subsequence. Note that 1536 is the windows size of the LED model, and 1024 is the maximum length supported by the BART model.

\begin{table*}
    \centering
    \begin{tabular}{@{\extracolsep{5pt}}l|lc@{\hspace{0.7\tabcolsep}}c@{\hspace{0.7\tabcolsep}}cc@{\hspace{0.7\tabcolsep}}c@{\hspace{0.7\tabcolsep}}cc@{\hspace{0.7\tabcolsep}}c@{\hspace{0.7\tabcolsep}}cc@{\hspace{0.7\tabcolsep}}c@{\hspace{0.7\tabcolsep}}c@{}}
    \thick
    &\multirow{2}{*}{Encoder} & \multicolumn{3}{c}{$L$=512} & \multicolumn{3}{c}{$L$=1024} & \multicolumn{3}{c}{$L$=1536} & \multicolumn{3}{c}{$L$=$\infty$}\\
    \cline{3-5} \cline{6-8} \cline{7-11} \cline{12-14}
    && R1 & R2 & RL & R1 & R2 & RL & R1 & R2 & RL & R1 & R2 & RL  \\
    \hline
\multirow{2}{*}{\rotatebox{90}{SF}}& BART  & 26.3&5.1&16.2&27.2&4.9&16.7&-&-&-&-&-&-\\
& LED  & 32.8&7.0&18.8&33.1&7.3&18.9&33.2&7.0&18.6&33.6&7.1&18.7\\
\hline
\multirow{2}{*}{\rotatebox{90}{GR}}& BART  & 45.6&16.9&21.8&47.9&18.6&22.7&-&-&-&-&-&-\\
& LED  & 53.9&24.7&27.1&54.1&25.1&27.9&54.8&25.7&27.8&56.6&26.6&29.1\\
    \thick
    \end{tabular}
    \caption{Full results on summarization. ``SS'' stands for the SummScreen dataset, and ``GR'' stands for the GovReport dataset. The BART model does not support sequence longer than 1024 tokens.}
    \label{tab:full_summ}
\end{table*}

\begin{table*}[t]
    \centering
    \begin{tabular}{@{\extracolsep{2pt}}lccccccccccc@{}}
    \thick
    \multirow{2}{*}{Encoder} & \multicolumn{3}{c}{MUC} & \multicolumn{3}{c}{B$^3$} & \multicolumn{3}{c}{CEAF$_{\phi_4}$} & \multirow{2}{*}{Avg.}\\
    \cline{2-4} \cline{5-7} \cline{8-10}
    & P & R & F1 & P & R & F1 & P & R & F1 & \\
    \hline
BigBird~{\small($L$=128)} & 84.5 & 78.5 & 81.4 & 75.2 & 68.6 & 71.7 & 73.5 & 63.2 & 68.0 & 73.7 \\
BigBird~{\small($L$=256)} & 85.1 & 80.3 & 82.6 & 76.7 & 71.2 & 73.8 & 74.7 & 66.4 & 70.3 & 75.6 \\
BigBird~{\small($L$=512)} & 85.7 & 82.3 & 84.0 & 77.9 & 73.9 & 75.9 & 75.7 & 69.4 & 72.4 & 77.4 \\
BigBird~{\small($L$=1024)} & 85.2 & 82.5 & 83.8 & 77.1 & 74.5 & 75.8 & 76.3 & 69.6 & 72.8 & 77.4 \\
BigBird~{\small($L$=4096)} & 85.1 & 82.8 & 83.9 & 77.7 & 75.1 & 76.4 & 75.6 & 70.1 & 72.7 & 77.7 \\
Longformer~{\small($L$=128)} & 84.4 & 80.0 & 82.1 & 75.3 & 70.5 & 72.8 & 72.9 & 66.0 & 69.3 & 74.8 \\
Longformer$^G$~{\small($L$=128)} & 84.4 & 79.1 & 81.7 & 75.1 & 69.2 & 72.0 & 72.8 & 65.2 & 68.8 & 74.2 \\
Longformer~{\small($L$=256)} & 84.8 & 81.7 & 83.2 & 76.2 & 72.5 & 74.3 & 73.9 & 68.8 & 71.2 & 76.3 \\
Longformer$^G$~{\small($L$=256)} & 84.4 & 81.8 & 83.1 & 75.6 & 73.1 & 74.3 & 74.3 & 68.3 & 71.2 & 76.2 \\
Longformer~{\small($L$=512)} & 85.5 & 82.6 & 84.0 & 78.0 & 74.5 & 76.2 & 75.2 & 70.7 & 72.9 & 77.7 \\
Longformer$^G$~{\small($L$=512)} & 84.5 & 83.4 & 83.9 & 76.2 & 75.2 & 75.7 & 74.3 & 70.9 & 72.6 & 77.4 \\
Longformer~{\small($L$=1024)} & 86.0 & 82.1 & 84.0 & 78.7 & 73.4 & 76.0 & 75.2 & 70.2 & 72.6 & 77.5 \\
Longformer$^G$~{\small($L$=1024)} & 82.4 & 79.2 & 80.8 & 72.2 & 69.5 & 70.8 & 72.2 & 65.4 & 68.6 & 73.4 \\
Longformer~{\small($L$=4096)} & 85.2 & 82.9 & 84.1 & 77.4 & 74.6 & 76.0 & 74.8 & 70.7 & 72.7 & 77.6 \\
RoBERTa~{\small($L$=128)} & 81.1 & 78.0 & 79.6 & 70.5 & 68.0 & 69.3 & 71.2 & 63.3 & 67.0 & 72.0 \\
RoBERTa$^p$~{\small($L$=128)} & 61.3 & 45.0 & 51.9 & 45.9 & 30.3 & 36.5 & 41.4 & 25.8 & 31.8 & 40.1 \\
RoBERTa~{\small($L$=256)} & 84.7 & 81.8 & 83.2 & 76.0 & 72.6 & 74.3 & 74.2 & 68.5 & 71.3 & 76.3 \\
RoBERTa$^p$~{\small($L$=256)} & 67.7 & 46.0 & 54.8 & 53.0 & 30.8 & 39.0 & 43.8 & 26.9 & 33.3 & 42.4 \\
RoBERTa~{\small($L$=512)} & 85.7 & 82.6 & 84.1 & 78.1 & 74.3 & 76.2 & 75.2 & 70.9 & 73.0 & 77.8 \\
RoBERTa$^p$~{\small($L$=512)} & 67.0 & 45.0 & 53.9 & 53.0 & 29.8 & 38.1 & 43.2 & 26.8 & 33.1 & 41.7 \\
SpanBERT~{\small($L$=128)} & 78.2 & 75.5 & 76.8 & 66.6 & 64.3 & 65.5 & 68.2 & 60.5 & 64.1 & 68.7 \\
SpanBERT$^p$~{\small($L$=128)} & 56.5 & 45.7 & 50.5 & 39.8 & 31.5 & 35.1 & 38.4 & 25.1 & 30.4 & 38.7 \\
SpanBERT~{\small($L$=256)} & 83.2 & 79.6 & 81.4 & 74.4 & 70.1 & 72.2 & 71.7 & 67.0 & 69.3 & 74.3 \\
SpanBERT$^p$~{\small($L$=256)} & 64.1 & 46.5 & 53.9 & 49.8 & 31.8 & 38.8 & 40.7 & 27.9 & 33.1 & 41.9 \\
SpanBERT~{\small($L$=512)} & 83.5 & 81.3 & 82.4 & 74.6 & 71.9 & 73.2 & 72.3 & 68.5 & 70.3 & 75.3 \\
SpanBERT$^p$~{\small($L$=512)} & 63.7 & 47.0 & 54.1 & 48.7 & 32.0 & 38.6 & 42.2 & 27.9 & 33.6 & 42.1 \\
XLNet~{\small ($L$=128, $m$=0)} & 79.4 & 39.8 & 53.0 & 69.1 & 30.1 & 41.9 & 61.6 & 32.7 & 42.7 & 45.9 \\
XLNet~{\small ($L$=128, $m$=128)} & 78.1 & 49.1 & 60.3 & 66.0 & 39.4 & 49.3 & 63.8 & 38.8 & 48.2 & 52.6 \\
XLNet~{\small ($L$=256, $m$=0)} & 78.8 & 59.5 & 67.8 & 68.0 & 48.6 & 56.7 & 66.4 & 47.9 & 55.7 & 60.1 \\
XLNet~{\small ($L$=256, $m$=256)} & 64.6 & 67.5 & 66.0 & 48.4 & 55.0 & 51.5 & 59.8 & 45.3 & 51.6 & 56.4 \\
XLNet~{\small ($L$=512, $m$=0)} & 80.3 & 71.7 & 75.7 & 70.7 & 61.4 & 65.7 & 66.9 & 60.0 & 63.3 & 68.2 \\
XLNet~{\small ($L$=512, $m$=512)} & 76.2 & 73.0 & 74.6 & 64.2 & 63.8 & 64.0 & 66.4 & 58.3 & 62.1 & 66.9 \\

\thick
    \end{tabular}
    \caption{Full results on Ontonotes with the start2end model. We adopt the same notation as we used in \cref{tab:detail_coref_c2f}.}
    \label{tab:detail_coref_s2e}
\end{table*}

\end{document}